%% file: main.tex
\documentclass[letterpaper]{article} 
\usepackage{aaai2026}  
\usepackage{times}  
\usepackage{helvet}  
\usepackage{courier}  
\usepackage[hyphens]{url}  
\usepackage{graphicx} 
\usepackage{natbib}  
\usepackage{caption} 
\usepackage{algorithm}
\usepackage{algorithmic}
\usepackage{graphicx}
\usepackage{amsmath}
\usepackage{amssymb}
\usepackage{booktabs}
\usepackage{multirow}
\usepackage{xcolor}
\usepackage{colortbl, booktabs}
\usepackage{subcaption}
\usepackage{array}
\usepackage{hhline}
\usepackage{comment}
\usepackage{siunitx}

\usepackage{newfloat}
\usepackage{listings}
\DeclareCaptionStyle{ruled}{labelfont=normalfont,labelsep=colon,strut=off} 
\floatstyle{ruled}
\newfloat{listing}{tb}{lst}{}
\floatname{listing}{Listing}
\title{UniMM-V2X: MoE-Enhanced Multi-Level Fusion for End-to-End Cooperative Autonomous Driving}
\author{
    Ziyi Song\textsuperscript{\rm 1},
    Chen Xia\textsuperscript{\rm 1},
    Chenbing Wang\textsuperscript{\rm 1}
    Haibao Yu\textsuperscript{\rm 2},
    Sheng Zhou\textsuperscript{\rm 1, 3}\thanks{Corresponding author.},
    Zhisheng Niu\textsuperscript{\rm 1}
}
\affiliations{

    \textsuperscript{\rm 1}Department of Electronic Engineering, Tsinghua University\\
    \textsuperscript{\rm 2}The University of Hong Kong\\
    \textsuperscript{\rm 3}State Key Laboratory of Intelligent Green Vehicle and Mobility, Tsinghua University\\
    \{songzy24,xiac23,wcb24\}@mails.tsinghua.edu.cn,
    yuhaibao94@gmail.com,\\
    sheng.zhou@tsinghua.edu.cn, niuzhs@tsinghua.edu.cn
}

\usepackage{bibentry}

\begin{document}

\maketitle

\begin{abstract}
Autonomous driving holds transformative potential but remains fundamentally constrained by the limited perception and isolated decision-making with standalone intelligence. While recent multi-agent approaches introduce cooperation, they often focus merely on perception-level tasks, overlooking the alignment with downstream planning and control, or fall short in leveraging the full capacity of the recent emerging end-to-end autonomous driving. In this paper, we present UniMM-V2X, a novel end-to-end multi-agent framework that enables hierarchical cooperation across perception, prediction, and planning. At the core of our framework is a multi-level fusion strategy that unifies perception and prediction cooperation, allowing agents to share queries and reason cooperatively for consistent and safe decision-making. To adapt to diverse downstream tasks and further enhance the quality of multi-level fusion, we incorporate a Mixture-of-Experts (MoE) architecture to dynamically enhance the BEV representations. We further extend MoE into the decoder to better capture diverse motion patterns. Extensive experiments on the DAIR-V2X dataset demonstrate our approach achieves state-of-the-art (SOTA) performance with a 39.7\% improvement in perception accuracy, a 7.2\% reduction in prediction error, and a 33.2\% improvement in planning performance compared with UniV2X, showcasing the strength of our MoE-enhanced multi-level cooperative paradigm.
\end{abstract}

\begin{links}
    \link{Code}{https://github.com/Souig/UniMM-V2X}
\end{links}

\section{Introduction}
Traditional autonomous driving pipelines, with their modular structure, suffer from error propagation and limited generalization. As \cite{bevformer} improved environmental perception through bird's-eye-view (BEV) representations, end-to-end autonomous driving has been widely studied in \cite{uniad, vad, sparsedrive}. Although end-to-end autonomous driving offers a solution by directly mapping raw sensor data to final control, this standalone-intelligence system is constrained by sensor range and and struggle with rare critical events and predicting other agents' intentions. Vehicle-to-Everything (V2X) communication emerges as a key enabler to overcome these limitations by facilitating real-time information exchange. 

 \begin{figure}[t]
    \centering
    \includegraphics[width=1.0\linewidth]{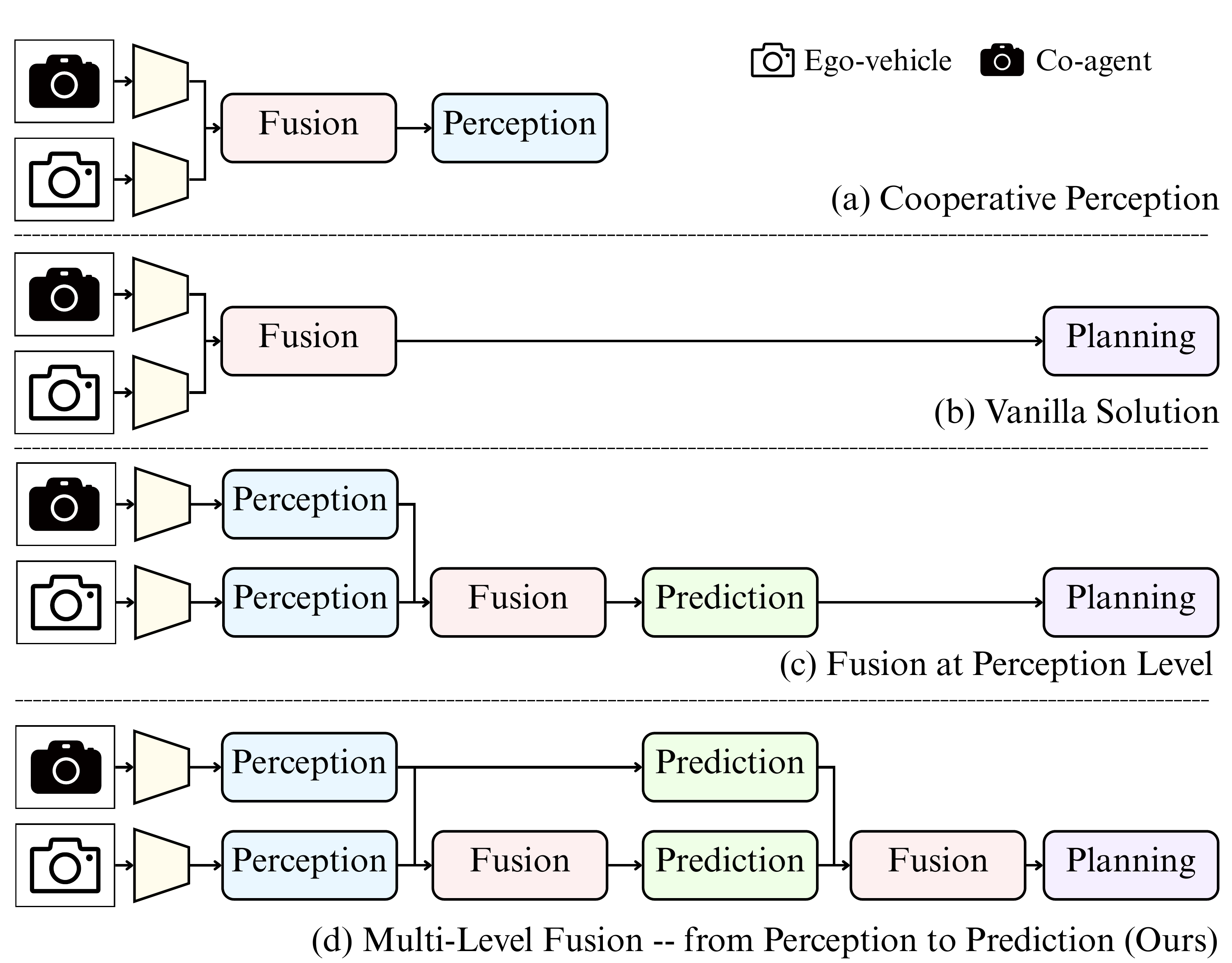}
    \caption{V2X communication modes in the VICAD (Vehicle-to-Infrastructure Cooperation Autonomous Driving) problem~\cite{dair}. (a) Cooperative perception methods focus on multi-agent detection and tracking, but may not align with planning objectives. (b) Vanilla solutions fuse features directly to generate planning outputs, with limited interpretability and compromised safety. (c) Module results can be supervised, but only enable perception-level cooperation. (d) Our design employs multi-level, multi-agent cooperation that integrates perception and prediction to enable cooperative decision-making.}
    \label{fig:v2x-mode}
\end{figure}

\begin{figure*}[t]
  \centering
   \includegraphics[width=1.0\linewidth]{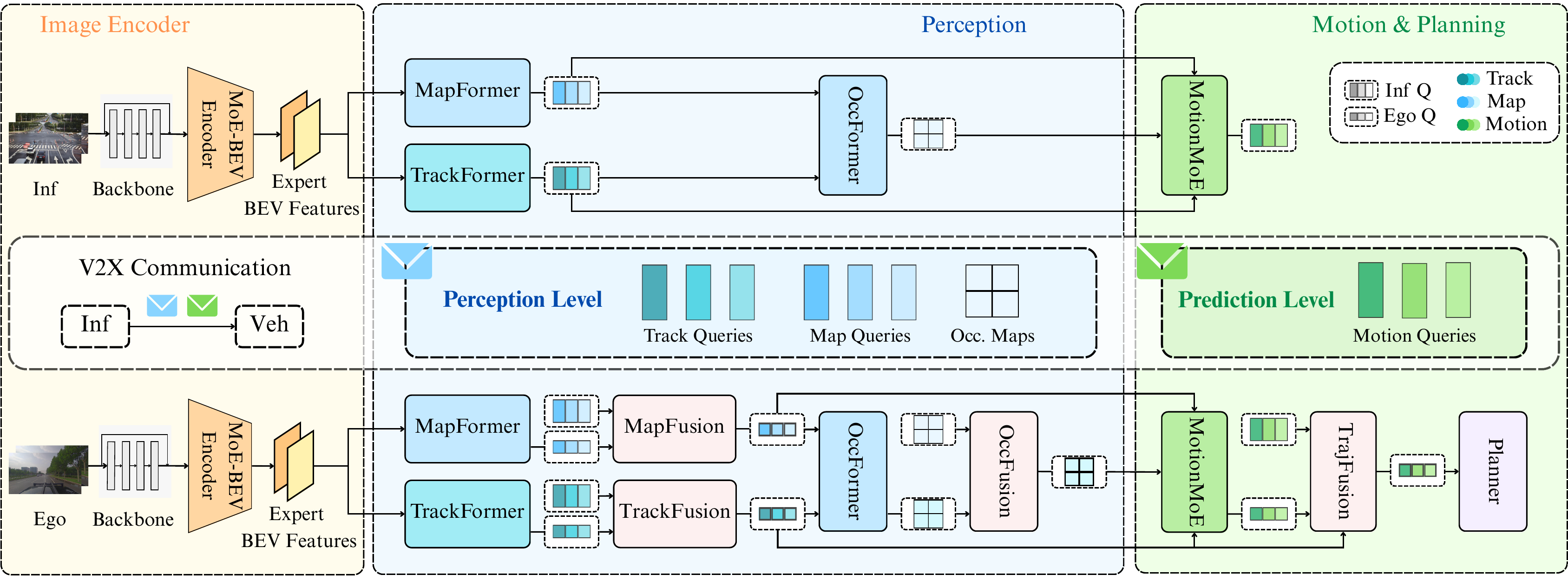}

   \caption{The overview of the UniMM-V2X framework. The system performs explicit multi-level fusion by integrating perception-level and prediction-level information from multiple agents to enhance downstream planning. Both the BEV encoder and motion decoder are equipped with MoE architectures, where the encoder generates task-adaptive BEV features tailored for various downstream tasks, and the decoder employs specialized experts to model diverse motion patterns, enhancing the effectiveness and adaptability of multi-level fusion for more robust planning performance. This unified MoE-enhanced multi-level fusion framework facilitates effective cooperation among agents throughout the entire autonomous driving pipeline.}
   \label{fig:framework}
\end{figure*}

As shown in Figure~\ref{fig:v2x-mode}(a), V2X communication is widely applied in cooperative perception, improving environmental awareness through multi-agent cooperation~\cite{v2x-vit, cobevt, fcooper, where2comm}. CooperNaut~\cite{coopernaut} encodes LiDAR into compact features for transmission but suffers from limited interpretability (Figure~\ref{fig:v2x-mode}(b)). UniV2X~\cite{univ2x} adopts a query-based architecture with sparse-dense hybrid communication but its fusion is restricted to the perception level (Figure~\ref{fig:v2x-mode}(c)). With end-to-end autonomous driving becoming the prevailing paradigm, a natural question arises: \emph{While multi-agent cooperation has improved perception, is this enough for end-to-end systems where planning is the final goal?}
 
Due to the complexity of end-to-end autonomous driving, relying solely on perception-level fusion is often insufficient, as accurate multi-agent motion prediction plays a more critical role in ensuring safety and efficiency. To address this, we propose UniMM-V2X, an MoE-enhanced \emph{multi-level} fusion framework that performs cooperative information fusion at both the perception and prediction levels, addressing the VICAD problem identified in \cite{univ2x}. At the perception level, we exchange track queries, map queries and occupancy probability maps to enable cooperative scene understanding. Building on this foundation, motion queries are further transmitted at the prediction level, allowing agents to reason jointly about future behaviors. Moreover, the interpretability of queries at both the instance and scene levels enhances the reliability of the system.

While multi-level fusion enables coherent information flow from perception to prediction, different downstream tasks have distinct requirements for BEV representations and other features. A shared BEV encoder may struggle to simultaneously meet the needs of perception, prediction, and planning, and the conventional motion decoder may fail to capture diverse agent motions. These concerns raise another natural question: \emph{How can the system adaptively generate specialized representations and predictions to meet these heterogeneous demands?}

To overcome these challenges, we innovatively integrate MoE architecture into both the \emph{BEV encoder} and \emph{motion decoder}. The MoE-enhanced encoder dynamically generates task-specialized BEV representations, allowing each task to leverage features best suited to its objectives. Meanwhile, the MoE-equipped decoder further dynamically generates motion queries via expert branches, each modeling distinct motion patterns such as keeping forward, turning left, or turning right. By complementing multi-level fusion, this design not only improves decision quality but also enhances interpretability and reliability.

The integration of multi-level fusion and MoE architecture creates a synergistic effect beyond individual gains. When combined, perception-level fusion benefits from more specialized BEV features generated by task-aware MoE encoders, and prediction-level fusion receives more reliable trajectory candidates guided by expert-decoded motion queries. This close integration enables each stage of the cooperative autonomous driving pipeline to perform more effectively and consistently, ultimately leading to more accurate and robust driving decisions.

\begin{figure*}
    \centering
   \includegraphics[width=0.9\linewidth]{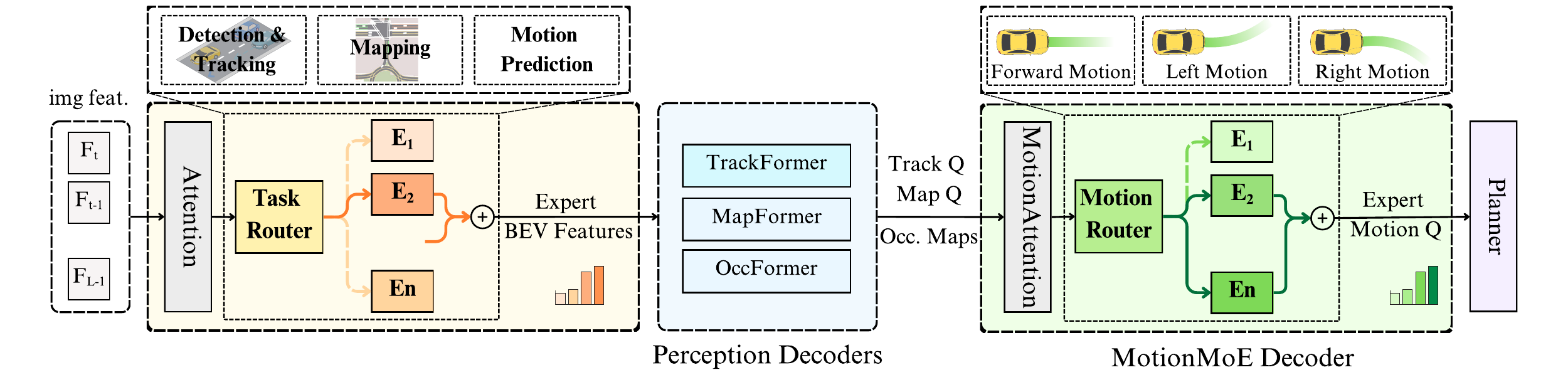}

   \caption{MoE-enhanced encoder and decoder in UniMM-V2X. The encoder enriches BEV feature extraction for diverse downstream tasks (e.g., detection, tracking, mapping, motion prediction), while the decoder generates motion queries through motion-specific experts (e.g., going forward, turning left, turning right) to improve planning quality.}
   \label{fig:moe}
\end{figure*}

Our contributions are summarized as follows:
\begin{itemize}
    \item To the best of our knowledge, we are the first to explore multi-level cooperation in multi-agent end-to-end autonomous driving, enabling cooperation across both perception and prediction to significantly improve decision-making reliability under complex scenarios.

    \item We introduce MoE into both the encoder and decoder of the end-to-end framework, enhancing the flexibility and specialization of the model to adapt to diverse tasks and prediction requirements in autonomous driving.

    \item Through extensive experiments, we validate that the combination of multi-level fusion and MoE architecture yields a strong complementary effect, facilitating more reliable cooperation and substantially improving decision quality. Compared to single-agent and existing cooperative methods, UniMM-V2X achieves SOTA results in perception, prediction, and planning.

\end{itemize}

\section{Related Works}
\subsection{End-to-End Autonomous Driving}
End-to-end autonomous driving has attracted growing attentions. Early methods~\cite{imitatelearning,behaviorclone,roach} lack interpretability and optimization by skipping intermediate tasks. Furthermore, ST-P3~\cite{st-p3} builds interpretable maps from perception; UniAD~\cite{uniad} unifies perception, prediction, and planning via a query-based framework; VAD~\cite{vad} and SparseDrive~\cite{sparsedrive} reduce computational cost through vectorization or sparse design; DiffusionDrive~\cite{diffusiondrive} employs diffusion models for planning. However, these methods are limited to single-agent inputs. In this work, we extend the paradigm to a multi-agent setting by incorporating cross-agent communication, joint perception-level and prediction-level fusion within a unified end-to-end framework for cooperative planning.

\subsection{Cooperative Autonomous Driving}
V2X communication in autonomous driving has its roots in early frameworks \cite{fcooper, cooper}. With the emergence of Transformer-based architectures, methods like \cite{who2com, where2comm, v2x-vit} have improved communication strategies by learning when, where and with whom to communicate. CooperNaut~\cite{coopernaut} goes further by linking perception and control into a unified framework. UniV2X~\cite{univ2x} introduces a sparse-dense hybrid communication protocol, effectively coordinating vehicle-to-infrastructure information. However, these methods either adopt vanilla fusion strategies or perform fusion only at the perception stage, limiting their effectiveness in planning. In contrast, we propose a multi-level fusion framework that operates across both perception and prediction stages, enabling agents to cooperatively reason from spatial observations to motion intents for safer planning.

\subsection{Mixture of Experts}
MoE operates using conditional computation and a learnable gating function. Early work, such as \cite{moe-mesh}, enabled MoE scaling in Transformers by replacing standard FFNs with sparsely activated experts, an idea later extended to large encoder-decoder models by \cite{gshard, moe-trans, st-moe} to address stability and fine-tuning issues. In autonomous driving, DriveMoE~\cite{drivemoe} applies MoE for sensor scheduling and action guidance. However, these methods typically restrict MoE to a single stage, limiting its ability to handle heterogeneous task demands. In contrast, we integrate MoE into both the BEV encoder and the motion decoder, enabling the system to generate task-specialized BEV representations as well as expert-guided motion predictions to improve decision reliability in multi-agent cooperation.

\begin{figure*}
    \centering
   \includegraphics[width=1.0\linewidth]{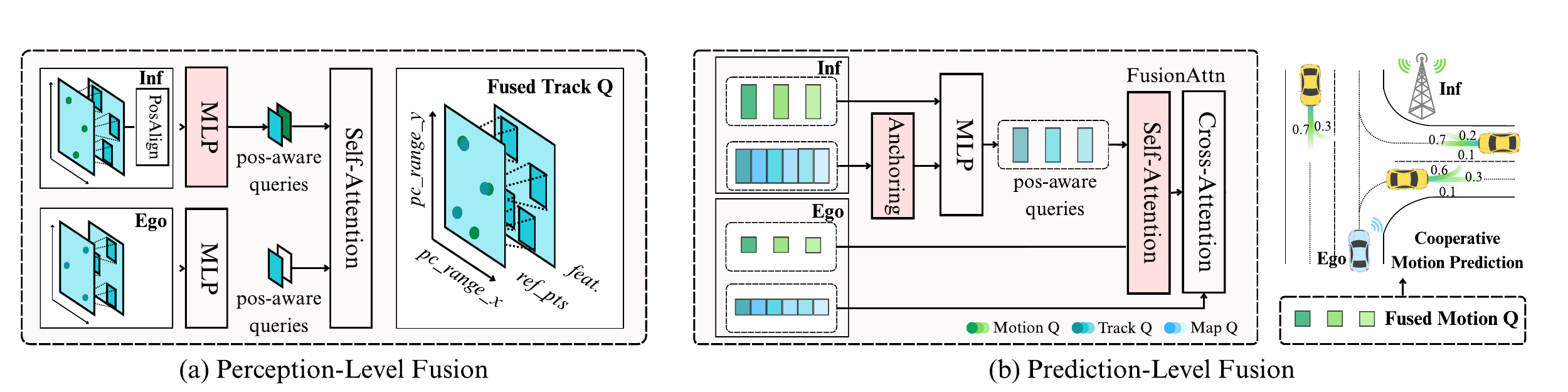}

   \caption{Multi-level fusion in UniMM-V2X. (a) Perception-level fusion introduces positional priors via reference point embeddings and uses attention-based dynamic fusion across agents. (b) Prediction-level fusion employs anchor-based embedding and dynamic fusion to support motion reasoning in complex multi-agent settings.}
   \label{fig:fusion}
\end{figure*}

\section{Method}
\subsection{Overview}
The overall framework of UniMM-V2X is illustrated in Figure~\ref{fig:framework}. It performs explicit \emph{multi-level} fusion across agents by integrating information at both the perception level and the prediction level, thereby enhancing the safety and robustness of downstream planning decisions. The MoE architecture is integrated into both the BEV encoder and the motion decoder, strengthening the effectiveness of multi-level fusion across perception and prediction. The encoder generates feature representations that are better adapted to the distinct needs of various downstream tasks, while the decoder exploits expert specialization to more precisely capture diverse motion patterns, ultimately delivering more robust and planning-aware trajectory outputs.

The framework consists of three main components: image encoders, a cooperative perception module, and a cooperative motion and planning module. The image encoder incorporates the MoE architecture to extract task-adaptive BEV features. The perception module performs cooperative detection, tracking, mapping, and occupancy map generation. The motion and planning module generates motion predictions via MoE-based decoder and fuses multi-agent predictions for planning decisions. Together, the perception-level and prediction-level fusion form a unified multi-level fusion framework that enables effective cooperation across agents throughout the decision-making process.

\subsection{MoE for Adaptive Feature and Motion Modeling}

To effectively address the complex joint demands of perception, prediction and planning, we place the MoE architecture in both the BEV encoder (MoE-BEV Encoder) and motion decoder (MotionMoE), as illustrated in Figure~\ref{fig:moe}. We adopt a standard sparse MoE design~\cite{moe, moe-trans}, in which traditional FFNs are replaced by a set of expert networks:
\begin{equation}
\text{MoE}(x) = \sum_{i \in \mathcal{I}_k(x)} \tilde{G}_i(x) \cdot f_i(x),
\end{equation}
where \( f_i \) is the \(i\)-th expert, \( \tilde{G}_i(x) \) the normalized routing weight, and \( \mathcal{I}_k(x) \) the selected experts. To avoid expert collapse and ensure balanced usage, we add a load balancing loss~\cite{moe-trans}:
\begin{equation}
\mathcal{L}_{\text{moe}} = \lambda \left( \mathrm{Var}(p) + \mathrm{Var}(l) \right),
\end{equation}
where \( p \) are the routing probabilities, \( l \) the expert loads, and \(\lambda\) a weighting factor, encouraging uniform expert activation.

Within the MoE-BEV encoder, we replace the FFNs with the MoE block to enable adaptive and specialized generation of BEV feature representations:
\begin{equation}
\mathbf{z}^{(l+1)} = \text{MoE}(\text{CrossAttn}(\text{SelfAttn}(\mathbf{z}^{(l)}))),
\end{equation}
where \( \mathbf{z}^{(l)} \) is the BEV feature representations at layer \( l \), and the MoE module selectively activates top-$k$ experts (e.g., \( k = 2 \)) to process the attended BEV features. Similarly, in the decoder, we replace the FFNs with the MoE architecture in the MotionMoE module to generate motion queries \( Q_M^{\text{veh}} \) and \( Q_M^{\text{other}} \) that adapt to diverse motion patterns.

\subsection{Multi-Agent Perception-Level Fusion}

In perception-level fusion, we incorporate track fusion, map fusion, and occupancy fusion. Among them, track fusion is particularly critical, because the resulting track queries function as crucial contributing inputs for downstream tasks. To enhance their quality, we introduce \textit{TrackFusion} module, which dynamically builds associations between agents, as shown in Figure~\ref{fig:fusion}(a). The map fusion and occupancy fusion modules are provided in the Appendix, where the former uses an MLP and the latter adopts a max operation.

In the TrackFusion block, an attention mechanism is employed to model complex inter-agent query relationships and perform weighted feature fusion based on learned relevance scores, overcoming the limitations of hard matching methods that rely on fixed distance thresholds in previous works. Initially, queries from other agents $Q_A^{\text{other}}$ are transformed into the ego-vehicle's coordinate system using an MLP:
\begin{equation}
    Q_A^{\text{other}}=\text{MLP}([Q_A^{\text{other}},\mathcal{R}]),
\end{equation}
where $\mathcal{R}$ is the rotation matrix. Subsequently, the reference point information $P_A^{\text{other}}$ and $P_A^{\text{veh}}$ are integrated as spatial contextual priors into the dynamic feature correlation learning process, as formulated below:
\begin{equation}
    Q_A={\rm MHSA}(X_A+\text{MLP}(P_A)),
  \label{eq:mhsa}
\end{equation}
\begin{equation}
    X_A=\text{Concat}(Q_A^{\text{veh}},Q_A^{\text{other}}),
\end{equation}
\begin{equation}
    P_A=\text{Concat}(P_A^{\text{veh}},P_A^{\text{other}}).
\end{equation}
We employ an MLP to embed the spatial coordinates of each agent into a learnable representation. These spatial embeddings are concatenated with agent-specific queries and jointly fed into a multi-head self-attention (MHSA) mechanism. This design allows the model to capture semantic dependencies across agents while incorporating their relative spatial positions, enabling context-aware and spatially sensitive feature fusion that enhances cooperative understanding.

\begin{table*}[t]
\small
  \centering
  \begin{tabular}{l|ccc>{\columncolor{gray!20}}c|ccc>{\columncolor{gray!20}}c|c}
    \toprule
    \multirow{2}{*}{Method} & \multicolumn{4}{c|}{L2 Error (m)$\downarrow$} & \multicolumn{4}{c|}{Collision Rate (\%)$\downarrow$} & Trans. Cost \\
     & 1\textit{s} & 2\textit{s} & 3\textit{s} & Avg. & 1\textit{s} & 2\textit{s} & 3\textit{s} & Avg. & (BPS)$\downarrow$ \\
    \midrule
    VAD*~\cite{vad} & 1.65 & 2.72 & 3.80 & 2.72 & 0.86 & 1.21 & 1.28 & 1.12 & - \\
    UniAD*~\cite{uniad} & 1.26 & 2.22 & 3.06 & 2.18 & 0.88 & 1.18 & 1.32 & 1.13 & - \\
    SparseDrive*~\cite{sparsedrive} & 1.02 & 1.69 & 2.37 & 1.69 & 0.46 & 1.23 & 1.28 & 0.99 & - \\
    \midrule
    Vanilla & 1.36 & 2.29 & 3.32 & 2.32 & 1.03 & 0.88 & 1.32 & 1.08 & 8.19$\times10^7$ \\
    V2VNet~\cite{v2vnet} & 1.96 & 2.37 & 3.41 & 2.58 & 0.74 & 0.88 & 1.03 & 0.88 & 8.19$\times10^7$ \\
    CooperNaut~\cite{coopernaut} & 2.69 & 4.07 & 5.50 & 4.09 & 1.18 & 1.32 & 1.76 & 1.42 & 8.19$\times10^7$ \\
    UniV2X~\cite{univ2x} & 1.45 & 2.19 & 3.04 & 2.23 & 0.15 & \textbf{0.15} & 0.44 & 0.25 & 8.09{\boldmath$\times10^5$} \\
    \midrule
    \textbf{UniMM-V2X} & \textbf{0.78} & \textbf{1.63} & \textbf{2.05} & \textbf{1.49} & \textbf{0.05} & \textbf{0.15} & \textbf{0.15} & \textbf{0.12} & 9.32{\boldmath$\times10^5$} \\
    \bottomrule
  \end{tabular}
  \caption{\textbf{Planning} performance. *: Single-agent no fusion method. We achieve improvements in reducing L2 error and collision rate, enhancing overall system safety.}
  \label{tab:planning}
\end{table*}

\begin{table*}[t]
  \small
  \centering
  \begin{tabular}{l|>{\columncolor{gray!20}}c|>{\columncolor{gray!20}}c|>{\columncolor{gray!20}}cc|c}
    \toprule
    \multirow{2}{*}{Method} & \cellcolor{white} Detection & \cellcolor{white} Tracking & \multicolumn{2}{c}{Mapping} & Trans. Cost \\ [\dimexpr 0.5ex + 0.5pt]
    
     & mAP$\uparrow$ & AMOTA$\uparrow$ & Lane (\%)$\uparrow$ & Crossing (\%)$\uparrow$ & (BPS)$\downarrow$\\ 
    \midrule
    UniAD*~\cite{uniad} & 0.181 & 0.197 & 13.3 & 8.7 & - \\
    SparseDrive*~\cite{sparsedrive} & 0.324 & 0.130 & - & 5.2 & - \\
    \midrule
    Early Fusion & 0.243 & 0.209 & 16.7 & 17.8 & 8.19$\times10^7$ \\
    Late Fusion & 0.236 & 0.263 & 13.4 & 9.1 & \textbf{6.60$\times10^2$} \\
    \midrule
    CoAlign$^\dagger$~\cite{coalign} & 0.261 & 0.234 & - & - & 8.19$\times10^7$ \\
    Where2comm$^\dagger$~\cite{where2comm} & 0.221 & 0.106 & - & - & 5.40$\times10^5$ \\
    CoBEVT$^\dagger$~\cite{cobevt} & 0.264 & 0.243 & 15.6 & 16.4 & 2.56$\times10^6$\\
    V2X-ViT$^\dagger$~\cite{v2x-vit} & 0.261 & 0.287 & - & - & 2.56$\times10^6$ \\
    \midrule
    UniV2X~\cite{univ2x} & 0.302 & 0.241 & 17.7 & 19.7 & 2.17$\times10^5$ \\
    \midrule
    \textbf{UniMM-V2X} & \textbf{0.422} & \textbf{0.427} & \textbf{17.9} & \textbf{20.3} & 2.17$\times10^5$\\
    \bottomrule
  \end{tabular}
  \caption{\textbf{Perception} performance. *: Single-agent no fusion method. $\dagger$: Cooperative perception methods. We significantly improve all performance metrics without increasing transmission cost.}
  \label{tab:perception}
\end{table*}

\subsection{Cross-View Prediction-Level Fusion}
In prediction-level fusion, as shown in Figure~\ref{fig:fusion}(b), we fuse motion queries from multiple agents through \textit{TrajFusion} module to enable cooperative motion prediction, which finally improves the performance of planning decisions.

The fusion process begins with other agents transmitting their motion queries \( Q_M^{\text{other}}\) to the ego agent via inter-agent communication. To spatially align the heterogeneous trajectory data, we first transform the agent-level anchors \( P_{\text{anchor}}\), derived from \( Q_A^{\text{other}} \), into the coordinate frame of the ego-vehicle using the rotation matrix \( \mathcal{R} \):
\begin{equation}
P_{M}^{\text{other}} = \text{MLP}([P_{\text{anchor}}, \mathcal{R}]).
\end{equation}
The transformed positional information is then projected through an MLP for position embedding:
\begin{equation}
\tilde{Q}_M^{\text{other}} = \text{MLP}([Q_M^{\text{other}}, P_{M}^{\text{other}}]).
\end{equation}
The ego-agent motion queries and the positionally enhanced queries from other agents are then concatenated and processed via an attention-based mechanism:
\begin{equation}
F_M = \text{Concat}(Q_M^{\text{veh}}, \tilde{Q}_M^{\text{other}}),
\end{equation}
\begin{equation}
Q_M = \text{MHCA}(\text{MHSA}(F_M), Q_A).
\label{eq:TrajFusionAttn}
\end{equation}
Here, MHSA captures inter-agent dependencies within combined motion queries \( F_M \), and MHCA integrates perception-aware context by attending to the fused perception queries \( Q_A \), which are historically enriched and semantically aligned, thereby providing strong priors for motion reasoning in complex multi-agent scenarios.

\begin{table*}
    \small
    \raisebox{\dimexpr\ht\strutbox-\height}{
    \begin{minipage}[t]{0.42\linewidth}
    \centering
      \begin{tabular}{l|>{\columncolor{gray!20}}c|c}
        \toprule
        Method & IoU-n (\%)$\uparrow$ & IoU-f (\%)$\uparrow$ \\
        \midrule
        UniAD*~\cite{uniad} & 16.3 & 13.1 \\
        \midrule
        UniV2X~\cite{univ2x} & 22.2 & \textbf{26.0} \\
        \midrule
        \textbf{UniMM-V2X} & \textbf{23.0} & 23.7 \\
        \bottomrule
      \end{tabular}
      \caption{\textbf{Occupancy prediction} performance. “n” and “f” denote near (30×30m) and far (50×50m) ranges. *: Single-agent no fusion method.}
      \label{tab:occupancy}
    \end{minipage}
    }
    \hfill
    \raisebox{\dimexpr\ht\strutbox-\height}{
    \begin{minipage}[t]{0.55\linewidth}
    \centering
    
      \begin{tabular}{l|>{\columncolor{gray!20}}ccc}
      \toprule
      Method & minADE (m)$\downarrow$ & minFDE (m)$\downarrow$ & MR$\downarrow$ \\
      \midrule
      UniAD*~\cite{uniad} & 0.78 & 0.82 & 0.21 \\
      SparseDrive*~\cite{sparsedrive} & 1.02 & 1.87 & 0.34 \\
      \midrule
      UniV2X~\cite{univ2x} & 0.69 & 0.74 & 0.17 \\
      \midrule
      \textbf{UniMM-V2X} & \textbf{0.64} & \textbf{0.69} & \textbf{0.13} \\
      \bottomrule
      \end{tabular}
      \caption{\textbf{Moion prediction} performance. *: Single-agent no fusion method.}
      \label{tab:motion}
    \end{minipage}
    }
\end{table*}

\begin{table*}
  \centering
  \small
  \begin{tabular}{cc|cc|>{\columncolor{gray!20}}c>{\columncolor{gray!20}}c|>{\columncolor{gray!20}}c|ccc>{\columncolor{gray!20}}c|>{\columncolor{gray!20}}c}
    \toprule
    \multicolumn{2}{c|}{Multi-Level Fusion} & \multicolumn{2}{c|}{MoE} & \multicolumn{2}{c|}{Perception} & \cellcolor{white}Motion Prediction & \multicolumn{4}{c|}{Planning L2 Error (m)} & \cellcolor{white}Coll. (\%) \\ [\dimexpr 0.5ex + 0.5pt]
    P-Level & M-Level & Enc. & Dec. & mAP$\uparrow$ & AMOTA$\uparrow$ & minADE (m)$\downarrow$ & 1\textit{s} & 2\textit{s} & 3\textit{s} & Avg.$\downarrow$ & Avg.$\downarrow$ \\
    \midrule
    - & - & - & - & 0.181 & 0.197 & 0.78 & 1.26 & 2.22 & 3.06 & 2.18 & 1.13 \\
    \midrule
    \checkmark & - & - & - & 0.352 & 0.328 & 0.69 & 1.09 & 2.25 & 2.75 & 2.03 & 0.68 \\
    - & \checkmark & - & - & 0.191 & 0.193 & 0.67 & 1.13 & 1.71 & 2.71 & 1.85 & 0.50 \\
    \checkmark & \checkmark & - & - & 0.351 & 0.328 & 0.66 & 1.14 & 1.76 & 2.71 & 1.87 & 0.47 \\
    \midrule
    - & - & \checkmark & - & 0.238 & 0.269 & 0.81 & 1.28 & 1.97 & 2.92 & 2.06 & 0.39 \\
    - & - & - & \checkmark & 0.179 & 0.198 & 0.78 & 1.24 & 1.84 & 2.98 & 2.02 & 0.54 \\
    - & - & \checkmark & \checkmark & 0.240 & 0.267 & 0.75 & 1.02 & 1.73 & 2.82 & 1.85 & 0.24 \\
    \midrule
    \checkmark & - & \checkmark & \checkmark & \textbf{0.427} & \textbf{0.427} & 0.74 & 0.91 & 1.78 & 2.47 & 1.72 & 0.40 \\
    - & \checkmark & \checkmark & \checkmark & 0.238 & 0.271 & \underline{0.65} & 0.96 & \textbf{1.53} & 2.08 & \underline{1.52} & \underline{0.15} \\
    \checkmark & \checkmark & \checkmark & \checkmark & \underline{0.422} & \textbf{0.427} & \textbf{0.64} & \textbf{0.78} & 1.63 & \textbf{2.05} & \textbf{1.49} & \textbf{0.12} \\
    \bottomrule
  \end{tabular}
  \caption{\textbf{Ablation study results.} We conduct experiments to evaluate the effectiveness of multi-level fusion and the MoE mechanism. P-Level and M-Level refer to the perception level fusion and motion prediction level fusion, while Enc. and Dec. indicate applying MoE to the BEV encoder and motion decoder, respectively.}
  \label{tab:ablation}
\end{table*}

\subsection{Learning}
The overall training objective is to jointly optimize multiple sub-tasks involved in end-to-end cooperative autonomous driving. Specifically, the loss function consists of six components: detection and tracking, online mapping, occupancy prediction, motion prediction, planning, and the auxiliary load balancing term introduced by the MoE module.
\begin{equation}
    \mathcal{L} = \mathcal{L}_{\text{track}} + \mathcal{L}_{\text{map}} + \mathcal{L}_{\text{occ}} + \mathcal{L}_{\text{mot}} + \mathcal{L}_{\text{plan}} + \mathcal{L}_{\text{moe}}.
\end{equation}
All components are jointly optimized in an end-to-end manner to achieve unified perception, prediction, and planning.

\section{Experiments}
\label{sec:exp}
\subsection{Experimental Settings}
The overall framework is trained with the DAIR-V2X dataset~\cite{dair}, which comprises approximately 100 scenes captured at 28 complex traffic intersections in the real world. We use the AdamW optimizer with a learning rate of $\num{1e-4}$ and a weight decay of 0.01. We train the perception stage for 40 epochs on 8 NVIDIA A800 GPUs, and subsequently perform end-to-end motion and planning training for 20 epochs using the same GPU setup. During training, the MoE layers select the top-2 experts for each token to balance specialization and computational efficiency. Evaluation metrics of each task are described in the Appendix. We also implement UniMM-V2X on the V2X-Sim dataset~\cite{v2x-sim}, a large-scale simulation benchmark with diverse traffic scenarios for cooperative autonomous driving, and results are provided in the Appendix.

\subsection{Main Results}
We compare UniMM-V2X with several single-agent end-to-end autonomous driving models~\cite{vad, uniad, sparsedrive} as well as multi-agent cooperative driving frameworks. For the cooperative baselines, we evaluate both cooperative perception methods~\cite{coalign, where2comm, cobevt, v2x-vit} and end-to-end cooperative driving approaches~\cite{coopernaut, univ2x}.

\textbf{Planning.} The planning results are summarized in Table~\ref{tab:planning}. UniMM-V2X achieves the lowest average L2 error of \textbf{1.49m}, reducing by \textbf{33.2\%} compared with UniV2X~\cite{univ2x}, outperforming all the baselines including advanced single-agent and existing cooperative methods. More importantly, UniMM-V2X demonstrates superior safety, attaining the lowest average collision rate of \textbf{0.12\%}, which represents a \textbf{52.0\%} reduction compared to UniV2X~\cite{univ2x}. Although our approach introduces slightly higher communication overhead due to the transmission of motion queries, the improvements in the planning performance clearly justify the additional cost.

\textbf{Perception.} Table~\ref{tab:perception} presents the performance of UniMM-V2X on perception tasks. Compared to the SOTA no-fusion baseline~\cite{sparsedrive}, our method achieves a \textbf{+0.098} improvement in mAP and a \textbf{+0.297} improvement in AMOTA, demonstrating the effectiveness of cooperation. Compared to the SOTA end-to-end cooperative driving framework~\cite{univ2x}, our method achieves an improvement of \textbf{39.7\%} in mAP and \textbf{77.2\%} in AMOTA, without introducing additional communication cost at the perception level. For occupancy tasks, as shown in Table~\ref{tab:occupancy}, UniMM-V2X improves IoU-n by \textbf{+0.8\%} compared to UniV2X~\cite{univ2x}. 

\textbf{Prediction.} The motion prediction results are shown in Table~\ref{tab:motion}. UniMM-V2X achieves the best performance with \textbf{0.64m} minADE, \textbf{0.69m} minFDE and \textbf{13.2\%} MissRate, reducing errors by \textbf{7.2\%} and \textbf{6.8\%} on minADE and minFDE respectively compared with  UniV2X~\cite{univ2x}. These improvements contribute significantly to the improvement of the final planning performance mentioned above.

\begin{figure*}
    \centering
    \begin{subfigure}[b]{0.36\textwidth}
        \centering
        \includegraphics[width=\textwidth]{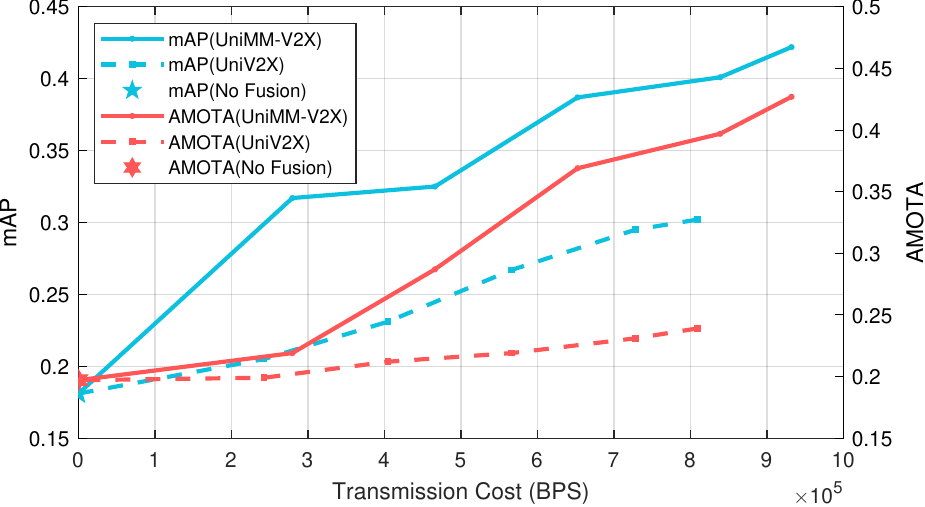}
        \caption{Perception performance.}
        \label{fig:communication-perception}
    \end{subfigure}
    \hspace{1em}
    \begin{subfigure}[b]{0.36\textwidth}
        \centering
        \includegraphics[width=\textwidth]{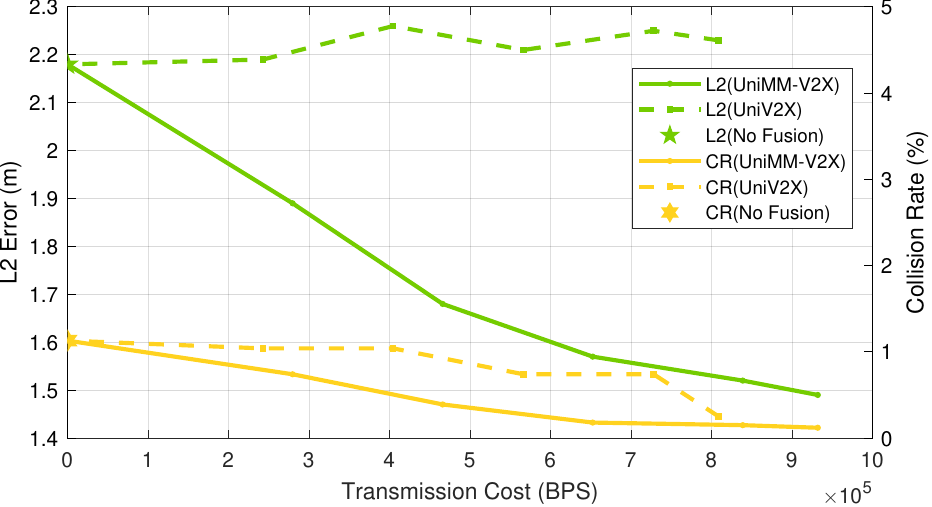}
        \caption{Planning performance.}
        \label{fig:communication-planning}
    \end{subfigure}
    
    \caption{Performance under different communication constraints.}
    \label{fig:communication}
\end{figure*}

\subsection{Ablation Study}

\textbf{Effect of Multi-Level Fusion}. As shown in Table~\ref{tab:ablation}, perception-level fusion improves detection and tracking performance but has limited effect on motion prediction and planning, probably due to the misalignment between perception accuracy and planning requirements. In contrast, prediction-level fusion enhances planning safety by providing supplementary motion cues for occluded objects and refining uncertain trajectories, but perception performance remains similar to the single-agent baseline due to the lack of early-stage cooperation. These observations indicate that single-level fusion alone is insufficient to optimize all driving tasks. Multi-level fusion ensures the propagation of high-quality intermediate features throughout the pipeline, resulting in consistent improvements across all modules.

\textbf{Effect of MoE.} As shown in Table~\ref{tab:ablation}, integrating MoE into the BEV encoder enhances environmental understanding, improving both perception and planning performance for the single vehicle. Using MoE only in the motion decoder yields limited gains, likely due to insufficient task-specific BEV features for accurate motion prediction. The best results are achieved when MoE is applied to both the encoder and the decoder, where the encoder produces task-aware BEV features and the decoder leverages expert specialization to capture complex motion behaviors.

\textbf{Interaction between Multi-Level Fusion and MoE.} Applying multi-level fusion individually does not achieve optimal performance, suggesting that traditional fusion can saturate when used independently at each level. In contrast, integrating MoE in both the encoder and decoder significantly enhances multi-level fusion, resulting in gains (AMOTA +0.230, L2 error -0.69m, Table~\ref{tab:ablation}). This indicates that multi-level fusion is crucial for capturing hierarchical interactions between perception and prediction, while MoE further enhances this capability by enabling task-specific specialization. Combined, they produce substantial improvements across perception, prediction, and planning.

\begin{table}[h]
  \centering
  \small
  \begin{tabular}{c|cc|c|c|c|c}
    \toprule
    Num. & P & M & mAP$\uparrow$ &AMOTA$\uparrow$ & L2 (m)$\downarrow$ & CR (\%)$\downarrow$ \\
    \midrule
    4 & - & - & 0.240 & 0.229 & 1.77 & 0.39 \\
    4 & \checkmark & - & 0.235 & 0.230 & 1.85 & 0.34 \\
    4 & - & \checkmark & 0.237 & 0.231 & 1.60 & 0.20 \\
    4 & \checkmark & \checkmark & 0.230 & 0.229 & 1.78 & 0.37 \\
    \midrule
    8 & - & - & \underline{0.421} & \underline{0.425} & \underline{1.66} & \underline{0.20} \\
    8 & \checkmark & - & 0.366 & 0.347 & 1.75 & 0.43 \\
    \rowcolor{gray!20}
    \textbf{8} & - & \textbf{\checkmark} & \textbf{0.422} & \textbf{0.427} & \textbf{1.49} & \textbf{0.12} \\
    8 & \checkmark & \checkmark & 0.368 & 0.351 & 1.68 & 0.25 \\
    \midrule
    16 & - & - & 0.403 & 0.374 & 1.71 & 0.36 \\
    16 & \checkmark & - & 0.359 & 0.341 & 1.76 & 0.41 \\
    16 & - & \checkmark & 0.401 & 0.374 & 1.64 & 0.15 \\
    16 & \checkmark & \checkmark & 0.361 & 0.339 & 1.72 & 0.20 \\
    \bottomrule
  \end{tabular}
  \caption{\textbf{Ablation on MoE expert number and decoder placement} conducted within the multi-level fusion framework. All the experiments utilize the MoE-based encoder. “P” indicates MoE applied to the perception decoder and “M” denotes its placement in the motion decoder.
}
  \label{tab:ablation-moe}
\end{table}

\textbf{MoE Configuration.} Since all decoder modules are Transformer-based, we explore replacing their FFNs with MoE. All variants employ the MoE-based encoder, where spatial experts consistently improve performance. Results in Table~\ref{tab:ablation-moe} show that using 8 experts in the motion decoder achieves the best trade-off. Too few experts limit the specialization, while too many experts cause data sparsity and under-utilization. Furthermore, limited token diversity in the perception decoder reduces MoE benefits and may introduce gating noise. In contrast, placing MoE in the motion decoder, which is more tightly coupled with the planner, enables better adaptation to diverse behaviors, leading to more flexible and accurate planning.

\subsection{Practicality and Reliability of the System} 
We assess the practicality and efficiency of our method by comparing communication cost and inference latency (BPS and FPS). Unlike bandwidth-heavy BEV methods, our query-based design drastically reduces communication cost by 87.9× without sacrificing planning quality. UniMM-V2X achieves an FPS of 5.4, which is a slight decrease compared to UniV2X's 5.8 FPS due to the integration of MoE and multi-level fusion, along with a modest increase in communication cost from enriched motion queries. However, these minor costs are strongly justified by significant improvements in planning safety and reliability, reflecting an excellent cost–benefit profile. Further evaluation under variable bandwidth conditions in Figure~\ref{fig:communication} shows that UniMM-V2X consistently outperforms UniV2X across all settings. While UniV2X's fusion offers negligible benefits to planning, performing close to the No Fusion baseline, our multi-level fusion with MoE approach can effectively leverage available communication for cooperative planning, ensuring reliability and scalability in real-world autonomous driving.

\section{Conclusion}
In this work, we propose UniMM-V2X, an end-to-end framework for robust multi-agent cooperative driving. By explicitly fusing information at both perception and prediction levels, and integrating MoE modules in the BEV encoder and motion decoder, the system adaptively handles diverse driving tasks and motion patterns. Extensive evaluations on the DAIR-V2X benchmark show that UniMM-V2X achieves state-of-the-art performance, with +39.7\% in detection, +77.2\% in tracking, -7.2\% motion prediction error, -33.2\% L2 error, and -52.0\% collision rate compared to previous SOTA method, while maintaining comparable communication cost. The framework demonstrates reliability under different bandwidth constraints, highlighting its practical deployability for real-world cooperative driving. Future work will extend UniMM-V2X to closed-loop evaluation, explore more communication-efficient fusion strategies, and further improve the robustness of multi-agent cooperation.

\section{Acknowledgments}
This work is sponsored in part by the project of Tsinghua University-Toyota Joint Research Center for AI Technology of Automated Vehicle.

\bibliography{aaai2026}

\newpage
\input{appendix}

\end{document}

%% file: appendix.tex
\lstset{%
	basicstyle={\footnotesize\ttfamily},
	numbers=left,numberstyle=\footnotesize,xleftmargin=2em,
	aboveskip=0pt,belowskip=0pt,%
	showstringspaces=false,tabsize=2,breaklines=true}
\floatstyle{ruled}
\newfloat{listing}{tb}{lst}{}
\floatname{listing}{Listing}
%
\pdfinfo{
/TemplateVersion (2026.1)
}

\setcounter{secnumdepth}{0} 

%


\title{Technical Appendix}



\twocolumn[\begin{center}
    \LARGE\bfseries Technical Appendix
\end{center} \vspace{0.35cm}]

\section{Implementation Details}

\paragraph{MoE Settings.} The architecture of both the encoder and decoder in our model consists of 6 layers of MoE, providing sufficient depth to capture spatial and temporal dependencies across multi-agent observations. Each MoE layer is equipped with 8 experts, employing top-2 expert routing to dynamically select relevant experts. Furthermore, we introduce a load balancing loss weight of $\lambda=0.03$ to ensure a balanced utilization of all experts. This setup enables adaptive specialization in processing tasks, improving the overall efficiency and performance of the model.

\paragraph{Perception.} The perception decoder is configured with 6 layers, and the perception range is set to a radius of 50 meters around the ego vehicle. We use a tracking threshold of 0.25, which is the minimum matching confidence required to associate objects across consecutive frames. A relatively low threshold ensures robustness when dealing with occlusion or partial views in challenging environments. 

For the perception-level fusion, we focus on three key components: TrackFusion, MapFusion and OccFusion. TrackFusion, which plays a critical role in downstream tasks, is detailed in the main text. For MapFusion, we fuse map queries using an MLP network:
\begin{equation}
    Q_L = \text{MLP}([Q_L^{\text{veh}}, \tilde{Q}_L^{\text{other}}]),
\end{equation}
where $\tilde{Q}_L^{\text{other}}$ is transformed into the coordinate frame of the ego vehicle. This transformation ensures that the map queries are aligned with the perspective of the ego vehicle. For OccFusion, aligned occupancy maps from different agents are combined using the following a max operation:
\begin{equation}
    \mathbf{P} = \text{max}(\mathbf{P}^{\text{veh}}, \tilde{\mathbf{P}}^{\text{other}}),
\end{equation}
\begin{equation}
    O(x, y) = 
    \begin{cases} 
    1, & \mathbf{P}(x, y) > \tau \\
    0, & \text{otherwise}
    \end{cases},
\end{equation}
where $\tilde{\mathbf{P}}^{\text{other}}$ is transformed into the view of the ego vehicle via a coordinate transformation, and $\tau=0.1$ is the threshold used for occupancy determination in our experiments.

\paragraph{Prediction.} In the prediction module, we define 6 motion modes per target, which allows the model to account for the uncertainty and multi-modality of future behavior. This enhances the ability of the model to predict a range of complex future trajectories. For trajectory-related tasks, the model uses 4 past steps and predicts the next 12 steps in the future.

\paragraph{Comparative Experiments.} To evaluate the effectiveness of our method, we conduct a series of comparative experiments across several fusion strategies and SOTA cooperative perception methods. We perform evaluations on the DAIR-V2X~\cite{dair} and V2X-Sim~\cite{v2x-sim} datasets, using the following fusion strategies:
\begin{itemize}
    \item \textbf{No Fusion:} No fusion of infrastructure data is performed. For this baseline, we compare against the UniAD~\cite{uniad} framework as well as the VAD~\cite{vad} and SparseDrive~\cite{sparsedrive} single-vehicle end-to-end SOTA algorithms to demonstrate the importance of multi-agent cooperation.
    \item \textbf{Early Fusion:} Raw data from the infrastructure is directly fused at the initial stage, following the setup from UniV2X~\cite{univ2x}.
    \item \textbf{Late Fusion:} Infrastructure perception results are fused using the Hungarian method~\cite{hungarian}.
    \item \textbf{Intermediate Fusion:} We reproduce current SOTA cooperative perception methods, including V2X-ViT~\cite{v2x-vit}, Where2comm~\cite{where2comm}, CoBEVT~\cite{cobevt}, and CoAlign~\cite{coalign}. For a fair comparison, we use only the image data from the agents and standardize evaluation settings.
\end{itemize}

For planning comparisons, we evaluate the performance of several representative baselines and fusion strategies. The vanilla method generates the final trajectory by simply fusing BEV features and passing them through an MLP to generate final planning results. For the V2VNet~\cite{v2vnet} and CooperNaut~\cite{coopernaut} methods, we use approaches similar to UniV2X~\cite{univ2x}, where the BEV features are fused and passed into the UniAD~\cite{uniad} framework for further processing. The results for UniV2X~\cite{univ2x} are directly reproduced from the official checkpoints for comparison.

\paragraph{Training.} Our training process is divided into four stages to ensure efficient learning of cooperative driving:
\begin{itemize}
    \item Stage 1: Pre-train the perception module of the infrastructure, including detection, tracking, and mapping.
    \item Stage 2: Pre-train the perception part of the ego vehicle.
    \item Stage 3: Introduce cooperation into the system and train the cooperative perception tasks between the ego vehicle and the infrastructure.
    \item Stage 4: Freeze the perception modules and focus on training the cooperative prediction and planning tasks, followed by fine-tuning to optimize overall performance.
\end{itemize}

\paragraph{Others.} The temporal query length is set to 5, meaning the model processes information from the past five frames. The perception range of the ego vehicle is $[-51.2m,-51.2m,51.2m,51.2m]$ while that of the infrastructure is $[0,-51.2m,102.4m,51.2m]$. We set the number of queries for tracking, mapping, and motion prediction to 1500, 300, and 500, respectively. The tracking queries ensure high recall in dense multi-object tracking scenarios, while the mapping and motion branches benefit from more focused query sets, optimizing their respective tasks.

\section{Evaluation Metrics}

We comprehensively evaluate our framework across six key tasks relevant to autonomous driving: object detection, multi-object tracking, online mapping, occupancy prediction, motion prediction, and planning, along with transmission cost to reflect communication overhead in the cooperative autonomous driving settings.

\paragraph{Object Detection and Multi-Object Tracking.} Following the standard NuScenes protocol~\cite{nuscenes}, we evaluate detection performance using mean Average Precision (mAP) based on the n-points interpolated precision-recall curve, which quantifies the average overlap between the predicted and ground-truth bounding boxes across different thresholds:
\begin{equation}
    \text{mAP} = \frac{1}{n-1} \sum_{r\in{\frac{1}{n-1},\frac{2}{n-1} ..., 1.0}} \text{AP}_r,
\end{equation}
\begin{equation}
    \text{AP}_r = \frac{\text{TP}_r}{\text{TP}_r + \text{FP}_r + \text{FN}_r},
\end{equation}
where $\text{TP}_r$, $\text{FP}_r$, and $\text{FN}_r$ represent true positives, false positives, and false negatives at IoU threshold $r$.

For tracking, we employ Average Multi-Object Tracking Accuracy (AMOTA), which captures both association accuracy and detection quality over recall levels:
\begin{equation}
\text{AMOTA} = \frac{1}{n-1} \sum_{r\in{\frac{1}{n-1},\frac{2}{n-1},...,1.0}} \text{MOTA}_r,
\end{equation}
\begin{equation}
\text{MOTA} = \text{max}(0,1 - \frac{\text{FN}_r + \text{FP}_r + \text{IDSW}_r-(1-r)\text{GT}}{r\text{GT}}),
\end{equation}
where $\text{IDSW}_r$ denotes identity switches, and $\text{GT}$ is the total number of ground-truth objects. Our detection and tracking experiments focus on the "Car" class, which is the most safety-critical in autonomous driving scenarios.

\paragraph{Online Mapping.} 
We assess the quality of high-definition map prediction using the IoU between predicted and ground-truth semantic map elements (e.g., lane markings, pedestrian crossings) from a BEV perspective:
\begin{equation}
\text{IoU} = \frac{|\text{Prediction} \cap \text{Ground Truth}|}{|\text{Prediction} \cup \text{Ground Truth}|}.
\end{equation}
This metric reflects the accuracy and completeness of online semantic map generation of the model.

\paragraph{Occupancy Prediction.} 
We follow UniAD~\cite{uniad} and measure scene-level semantic segmentation performance using IoU across two spatial ranges: near ($30 \times 30$m) and far ($50 \times 50$m). The metric evaluates the model's ability to differentiate between occupied and free space covering both static infrastructure (e.g., buildings, sidewalks) and dynamic agents (e.g., vehicles, pedestrians).

\paragraph{Motion Prediction.} 
We evaluate future trajectory forecasting using the following metrics:
\begin{itemize}
    \item Minimum Average Displacement Error (minADE): average $\ell_2$ distance between predicted and ground-truth points over the best-matching trajectory.
    \item Minimum Final Displacement Error (minFDE): $\ell_2$ distance at the final prediction timestamp.
    \item Miss Rate (MR): percentage of ground-truth trajectories that deviate beyond a threshold $\delta$ (typically 2m) from all predicted modes.
\end{itemize}
Formally, for $K$ predicted trajectories $\hat{Y}^k$ and a ground-truth $Y$, we compute:
\begin{equation}
\text{minADE} = \min_{k} \frac{1}{T} \sum_{t=1}^{T} \| \hat{Y}^k_t - Y_t \|_2,
\end{equation}
\begin{equation}
\text{minFDE} = \min_{k} \| \hat{Y}^k_T - Y_T \|_2.
\end{equation}

\paragraph{Planning.} 
We assess planning accuracy and safety using two primary metrics:
\begin{itemize}
    \item \textbf{L2 Error:} average $\ell_2$ distance between predicted and expert trajectories over time.
    \item \textbf{Collision Rate:} the percentage of predicted trajectories that result in collisions with obstacles or other agents.
\end{itemize}
We report these metrics at $1s$, $2s$, and $3s$ prediction horizons to reflect both short-term and long-term planning quality. The settings are the same as in ST-P3~\cite{st-p3}.

\paragraph{Transmission Cost.} 
To evaluate communication efficiency, we measure the bandwidth consumption using Bytes Per Second (BPS). Following the protocol of UniV2X~\cite{univ2x}, we assume that the communication frequency is 2 Hz and that all transmitted data including features and queries are serialized using 32-bit floats. Calibration matrices and ego-pose data are excluded from transmission cost computation. BPS reflects the total bandwidth required to support cooperation, and enables a trade-off analysis between performance and communication overhead. The slight increase in communication overhead of UniMM-V2X compared to UniV2X~\cite{univ2x} is due to the introduction of motion prediction level fusion.

\begin{table*}[t!]
  \centering
  \begin{tabular}{l|cccc|cccc}
    \toprule
    \multirow{2}{*}{Method} & \multicolumn{4}{c|}{L2 Error (m)$\downarrow$} & \multicolumn{4}{c}{Collision Rate (\%)$\downarrow$} \\
     & 1\textit{s} & 2\textit{s} & 3\textit{s} & Avg. & 1\textit{s} & 2\textit{s} & 3\textit{s} & Avg. \\
    \midrule
    No Fusion & 2.87 & 3.66 & 4.78 & 3.77 & 1.22 & 1.33 & 1.33 & 1.29 \\
    UniV2X & 2.44 & 2.92 & 3.90 & 3.09 & 1.00 & 0.88 & \textbf{0.77} & 0.88 \\
    \midrule
    \textbf{UniMM-V2X} & \textbf{2.00} & \textbf{2.22} & \textbf{3.16} & \textbf{2.46} & \textbf{0.66} & \textbf{0.77} & 1.00 & \textbf{0.81} \\
    \bottomrule
  \end{tabular}
  \caption{Planning performance on V2X-Sim dataset~\cite{v2x-sim}.}
  \label{tab:v2xsim-planning}
\end{table*}

\section{Results on V2X-Sim Dataset}
We also implement our proposed UniMM-V2X framework on the V2X-Sim dataset~\cite{v2x-sim}, which is crucial to evaluate the robustness of our model across various simulated cooperative driving scenarios. The V2X-Sim dataset offers a valuable benchmark to test the performance of our approach in a simulated environment with multiple cooperative agents.

\textbf{V2X-Sim Dataset.} The V2X-Sim dataset is a large-scale simulated dataset built on the CARLA platform, specifically designed for V2X-based cooperative autonomous driving research. It consists of 100 diverse urban driving scenes, which include a range of driving environments with varying levels of complexity and dynamic scenarios. Each scene features multiple connected vehicles equipped with multi-view cameras, LiDAR sensors, and high-definition (HD) maps. The dataset provides rich annotations for tasks such as object detection, multi-object tracking, motion prediction, and planning. These characteristics make V2X-Sim an ideal dataset for evaluating end-to-end multi-agent cooperation and assessing the effectiveness of different cooperative strategies in realistic driving contexts.

\begin{table}[h]
  \centering
  \small
  \begin{tabular}{l|cc}
    \toprule
    Method & minADE (m) $\downarrow$ & minFDE (m) $\downarrow$\\
    \midrule
    No Fusion & 0.79 & 0.97\\
    UniV2X~\cite{univ2x} & 0.76 & 0.92 \\
    \midrule
    \textbf{UniMM-V2X} & \textbf{0.64} & \textbf{0.75} \\
    \bottomrule
  \end{tabular}
  \caption{Motion prediction performance on the V2X-Sim dataset~\cite{v2x-sim}.}
  \label{tab:v2xsim-motion}
\end{table}

\textbf{Experiment Settings and Results.} In our experiments, we select two vehicles from the V2X-Sim dataset: one is designated as the ego vehicle, and the other as the cooperating agent. To ensure a fair comparison with existing methods, we use only the front-view camera data.

For training, we use a total of 80 scenes from the V2X-Sim dataset, ensuring a diverse range of driving environments. We use 10 scenes for validation and another 10 scenes for testing to evaluate the generalization ability of the model across different scenarios. In the original UniV2X~\cite{univ2x} framework, the planning evaluation follows a non-standard timing protocol. To maintain consistency and fairness in our evaluation, we adopt the standard evaluation protocol at $1s$, $2s$, and $3s$ intervals, which is widely used in the field to assess the accuracy and safety of autonomous driving systems.

The performance results for planning and motion prediction tasks are summarized in Table~\ref{tab:v2xsim-planning} and Table~\ref{tab:v2xsim-motion}. These results show that, compared to the SOTA end-to-end multi-agent cooperative approach UniV2X~\cite{univ2x}, our UniMM-V2X framework achieves significant improvements in key performance metrics. Specifically, our model reduces the L2 error by \textbf{20.39\%}, indicating a marked improvement in the accuracy of the planned trajectories. Additionally, the collision rate is lowered by \textbf{7.95\%}, which demonstrates a considerable enhancement in driving safety by minimizing risky behaviors in complex driving scenarios.

Furthermore, in terms of motion prediction, we observe a \textbf{15.8\%} reduction in the motion prediction error, which is crucial for improving planning performance. More accurate motion predictions provide the planner with better estimates of future trajectories, thereby enabling more informed and precise decision-making in dynamic and complex driving environments. This substantial improvement in prediction accuracy directly contributes to the overall safety and efficiency of the cooperative driving system, reinforcing the importance of accurate multi-agent cooperation.

\begin{table}[h]
  \small
  \centering
  \begin{tabular}{l|c|c|c}
    \toprule
    Method & Mem. (MB)$\downarrow$ & FPS $\uparrow$ & Trans. (MB)$\downarrow$ \\
    \midrule
    UniV2X & 6301 & 5.86 & 8.09$\times10^5$ \\
    \midrule
    UniMM-V2X & 6483 & 5.39 & 9.32$\times10^5$ \\
    
    \bottomrule
  \end{tabular}
  \caption{Inference complexity and transmission cost.}
  \label{tab:mem}
\end{table}

\section{Efficiency of the System} 

To assess the practicality and efficiency of our proposed method, we compare the inference complexity and communication cost with UniV2X~\cite{univ2x}, as summarized in Table~\ref{tab:mem}. Specifically, we report GPU memory usage (Mem.), processing speed in frames per second (FPS), and the total amount of data transmitted across agents (Trans.). These metrics are critical in understanding the trade-offs between computational efficiency and model performance in real-time multi-agent cooperative driving scenarios.

Overall, UniMM-V2X maintains competitive runtime performance while incurring only moderate increases in memory and communication costs. Compared to the baseline UniV2X, our full model increases GPU memory usage by approximately 2.9\% and reduces FPS by 0.47, primarily due to the integration of the MoE architecture and multi-level fusion strategy. Given that sparse MoE is often associated with improved computational efficiency, this increase in inference cost may appear counterintuitive. In contrast to large-scale vision-language models (VLMs), where MoE modules significantly reduce per-token computation by distributing workloads across many experts, our model size does not yet fully leverage such computational benefits, leading to a less pronounced efficiency gain.

Nevertheless, the introduction of MoE significantly enhances representation learning for both perception and prediction tasks, contributing to consistent improvements in downstream performance. Although communication overhead slightly increases due to more informative features and enriched motion queries being exchanged among agents, this cost is well justified. The resulting gains in prediction robustness and planning safety highlight the effectiveness of enhanced inter-agent cooperation. Therefore, UniMM-V2X achieves a favorable trade-off, achieving an effective balance between performance, efficiency, and scalability for real-time multi-agent autonomous driving systems.

\section{Qualitative Visualization}
\begin{figure*}
    \centering
    \begin{subfigure}[b]{0.75\textwidth}
        \centering
        \includegraphics[width=\textwidth]{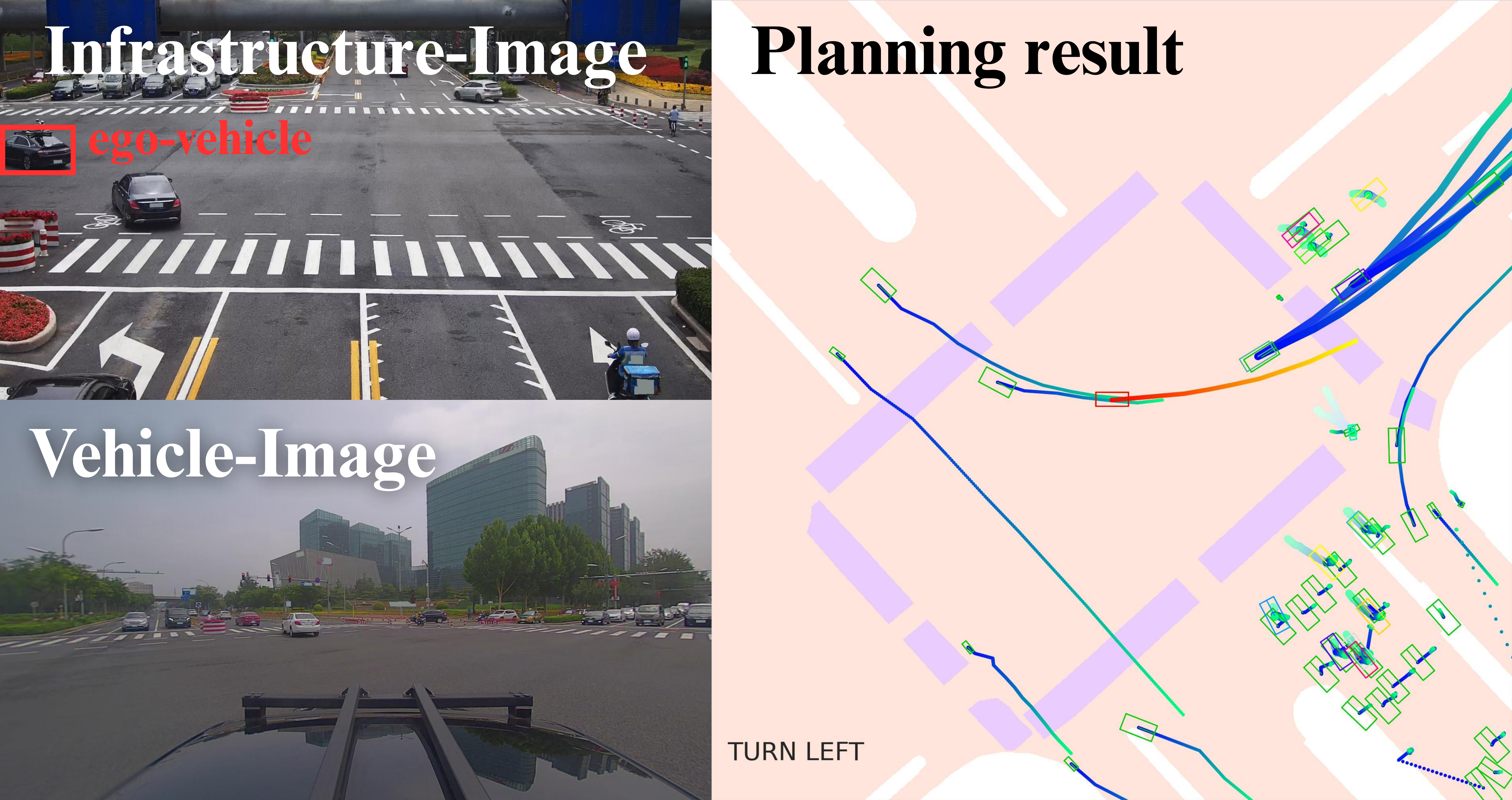}
        \caption{Turn left.}
        \label{fig:subimga}
    \end{subfigure}
    \begin{subfigure}[b]{0.75\textwidth}
        \centering
        \includegraphics[width=\textwidth]{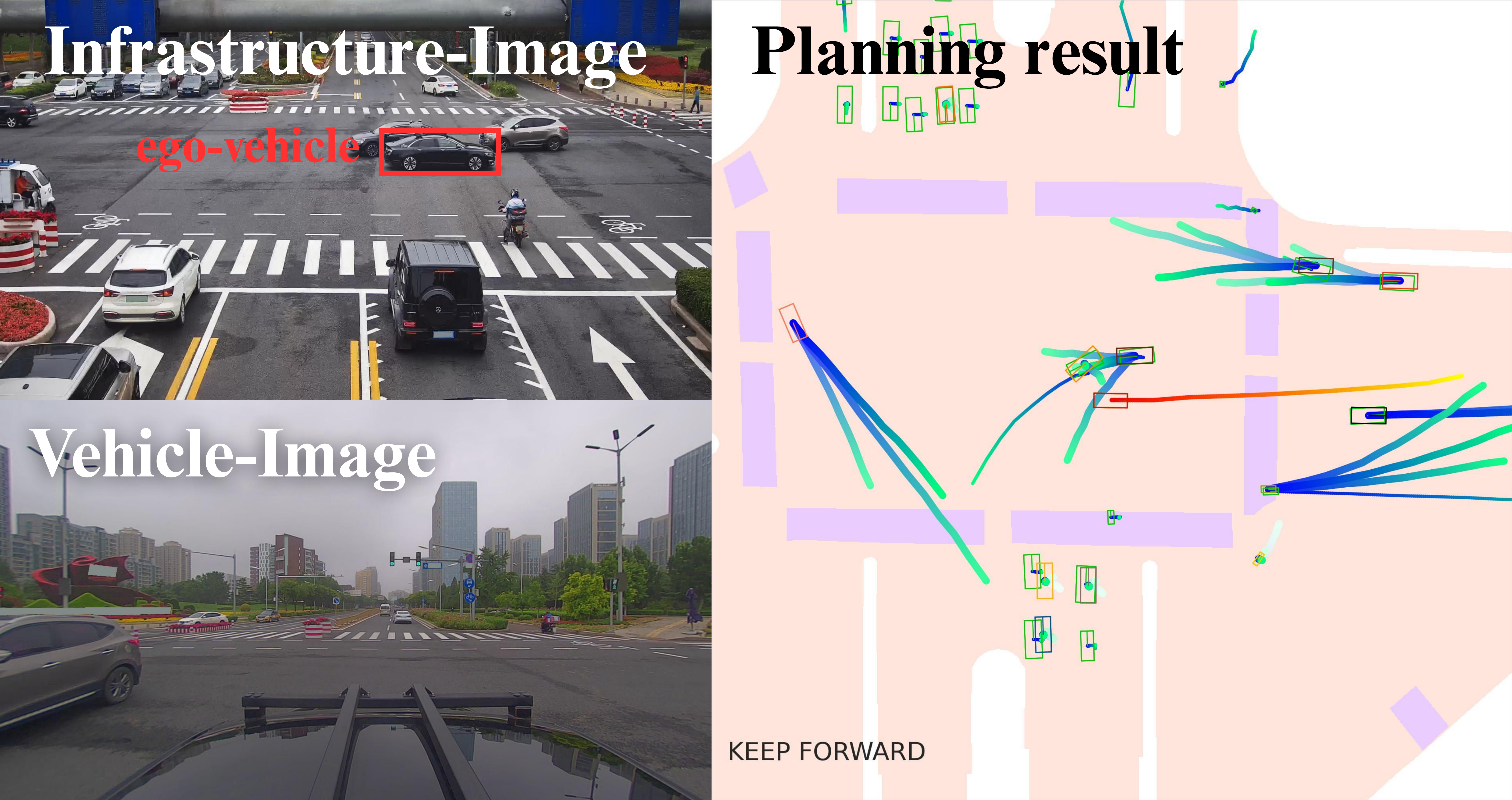} 
        \caption{Keep forward.}
        \label{fig:subimga}
    \end{subfigure}
    \begin{subfigure}[b]{0.75\textwidth}
        \centering
        \includegraphics[width=\textwidth]{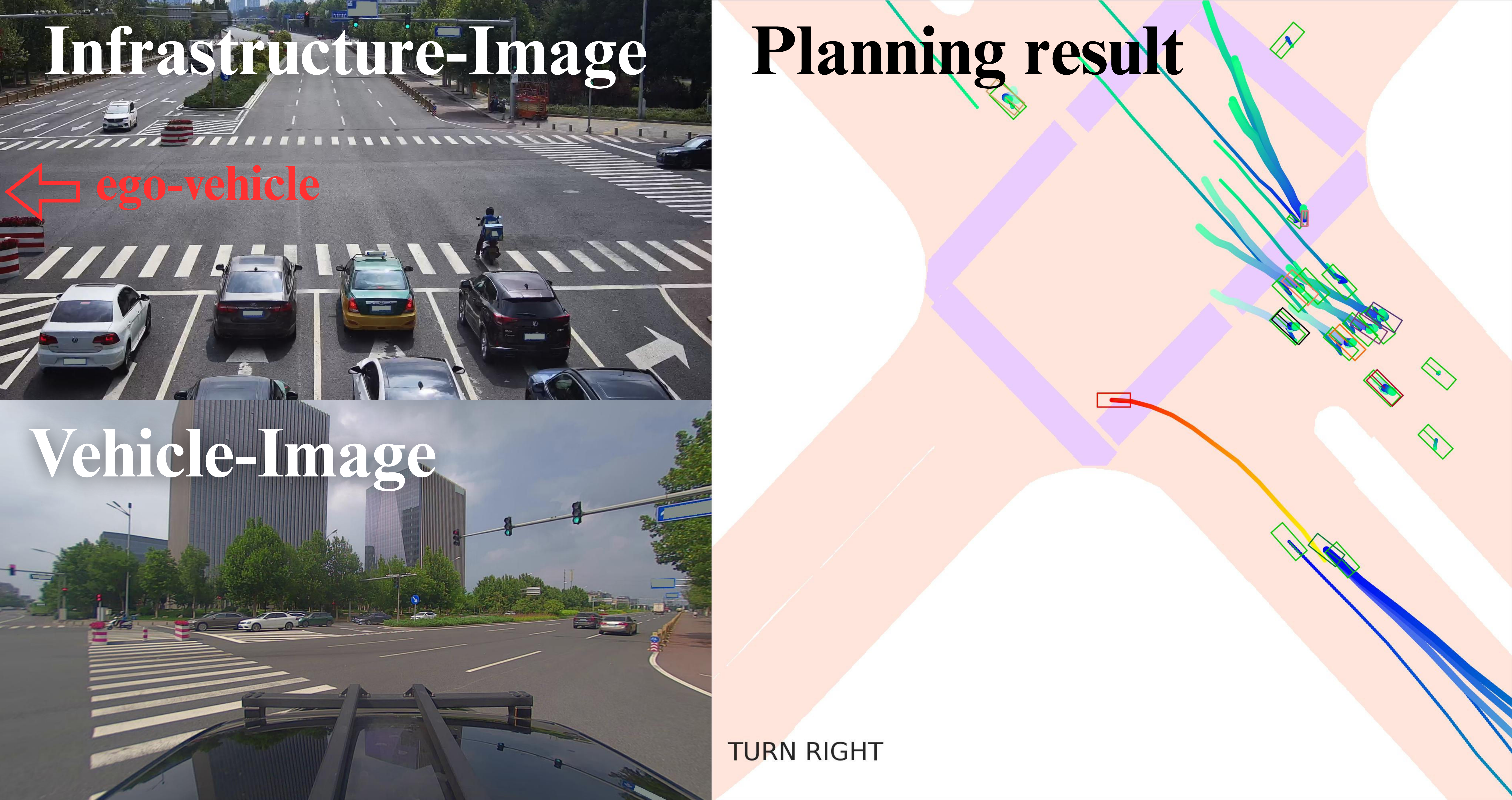} 
        \caption{Turn right.}
        \label{fig:subimga}
    \end{subfigure}
    \caption{UniMM-V2X’s planning performance on the DAIR-V2X dataset~\cite{univ2x}.}
    \label{fig:results}
\end{figure*}

We present qualitative results to illustrate the effectiveness of our MoE-enhanced multi-level end-to-end cooperative autonomous driving framework UniMM-V2X on the final planning performance. As shown in Figure~\ref{fig:results}, the visualizations cover different agent behaviors, including left turns, straight driving, and right turns, highlighting the excellent performance of UniMM-V2X in enhancing perception, prediction, and final planning.

%% file: aaai2026.bib
@inproceedings{uniad,
  title={Planning-oriented autonomous driving},
  author={Hu, Yihan and Yang, Jiazhi and Chen, Li and Li, Keyu and Sima, Chonghao and Zhu, Xizhou and Chai, Siqi and Du, Senyao and Lin, Tianwei and Wang, Wenhai and others},
  booktitle={Proceedings of the IEEE/CVF conference on computer vision and pattern recognition},
  pages={17853--17862},
  year={2023}
}

@inproceedings{vad,
  title={Vad: Vectorized scene representation for efficient autonomous driving},
  author={Jiang, Bo and Chen, Shaoyu and Xu, Qing and Liao, Bencheng and Chen, Jiajie and Zhou, Helong and Zhang, Qian and Liu, Wenyu and Huang, Chang and Wang, Xinggang},
  booktitle={Proceedings of the IEEE/CVF International Conference on Computer Vision},
  pages={8340--8350},
  year={2023}
}

@article{bevformer,
  title={Bevformer: learning bird's-eye-view representation from lidar-camera via spatiotemporal transformers},
  author={Li, Zhiqi and Wang, Wenhai and Li, Hongyang and Xie, Enze and Sima, Chonghao and Lu, Tong and Yu, Qiao and Dai, Jifeng},
  journal={IEEE Transactions on Pattern Analysis and Machine Intelligence},
  year={2024},
  publisher={IEEE}
}

@inproceedings{univ2x,
  title={End-to-end autonomous driving through v2x cooperation},
  author={Yu, Haibao and Yang, Wenxian and Zhong, Jiaru and Yang, Zhenwei and Fan, Siqi and Luo, Ping and Nie, Zaiqing},
  booktitle={Proceedings of the AAAI Conference on Artificial Intelligence},
  volume={39},
  number={9},
  pages={9598--9606},
  year={2025}
}

@inproceedings{v2x-vit,
  title={V2x-vit: Vehicle-to-everything cooperative perception with vision transformer},
  author={Xu, Runsheng and Xiang, Hao and Tu, Zhengzhong and Xia, Xin and Yang, Ming-Hsuan and Ma, Jiaqi},
  booktitle={European conference on computer vision},
  pages={107--124},
  year={2022},
  organization={Springer}
}

@article{cobevt,
  title={CoBEVT: Cooperative bird's eye view semantic segmentation with sparse transformers},
  author={Xu, Runsheng and Tu, Zhengzhong and Xiang, Hao and Shao, Wei and Zhou, Bolei and Ma, Jiaqi},
  journal={arXiv preprint arXiv:2207.02202},
  year={2022}
}

@inproceedings{fcooper,
  title={F-cooper: Feature based cooperative perception for autonomous vehicle edge computing system using 3D point clouds},
  author={Chen, Qi and Ma, Xu and Tang, Sihai and Guo, Jingda and Yang, Qing and Fu, Song},
  booktitle={Proceedings of the 4th ACM/IEEE Symposium on Edge Computing},
  pages={88--100},
  year={2019}
}

@inproceedings{v2vnet,
  title={V2vnet: Vehicle-to-vehicle communication for joint perception and prediction},
  author={Wang, Tsun-Hsuan and Manivasagam, Sivabalan and Liang, Ming and Yang, Bin and Zeng, Wenyuan and Urtasun, Raquel},
  booktitle={Computer vision--ECCV 2020: 16th European conference, Glasgow, UK, August 23--28, 2020, proceedings, part II 16},
  pages={605--621},
  year={2020},
  organization={Springer}
}

@article{where2comm,
  title={Where2comm: Communication-efficient collaborative perception via spatial confidence maps},
  author={Hu, Yue and Fang, Shaoheng and Lei, Zixing and Zhong, Yiqi and Chen, Siheng},
  journal={Advances in neural information processing systems},
  volume={35},
  pages={4874--4886},
  year={2022}
}

@inproceedings{coopernaut,
  title={Coopernaut: End-to-end driving with cooperative perception for networked vehicles},
  author={Cui, Jiaxun and Qiu, Hang and Chen, Dian and Stone, Peter and Zhu, Yuke},
  booktitle={Proceedings of the IEEE/CVF Conference on Computer Vision and Pattern Recognition},
  pages={17252--17262},
  year={2022}
}

@article{sparsedrive,
  title={Sparsedrive: End-to-end autonomous driving via sparse scene representation},
  author={Sun, Wenchao and Lin, Xuewu and Shi, Yining and Zhang, Chuang and Wu, Haoran and Zheng, Sifa},
  journal={arXiv preprint arXiv:2405.19620},
  year={2024}
}

@inproceedings{dair,
  title={Dair-v2x: A large-scale dataset for vehicle-infrastructure cooperative 3d object detection},
  author={Yu, Haibao and Luo, Yizhen and Shu, Mao and Huo, Yiyi and Yang, Zebang and Shi, Yifeng and Guo, Zhenglong and Li, Hanyu and Hu, Xing and Yuan, Jirui and others},
  booktitle={Proceedings of the IEEE/CVF Conference on Computer Vision and Pattern Recognition},
  pages={21361--21370},
  year={2022}
}

@article{moe-trans,
  title={Switch transformers: Scaling to trillion parameter models with simple and efficient sparsity},
  author={Fedus, William and Zoph, Barret and Shazeer, Noam},
  journal={Journal of Machine Learning Research},
  volume={23},
  number={120},
  pages={1--39},
  year={2022}
}

@article{moe,
  title={Outrageously large neural networks: The sparsely-gated mixture-of-experts layer},
  author={Shazeer, Noam and Mirhoseini, Azalia and Maziarz, Krzysztof and Davis, Andy and Le, Quoc and Hinton, Geoffrey and Dean, Jeff},
  journal={arXiv preprint arXiv:1701.06538},
  year={2017}
}

@misc{drivemoe,
      title={DriveMoE: Mixture-of-Experts for Vision-Language-Action Model in End-to-End Autonomous Driving}, 
      author={Zhenjie Yang and Yilin Chai and Xiaosong Jia and Qifeng Li and Yuqian Shao and Xuekai Zhu and Haisheng Su and Junchi Yan},
      year={2025},
      eprint={2505.16278},
      archivePrefix={arXiv},
      primaryClass={cs.CV},
      url={https://arxiv.org/abs/2505.16278}, 
}

@inproceedings{imitatelearning,
  title={End-to-end driving via conditional imitation learning},
  author={Codevilla, Felipe and M{\"u}ller, Matthias and L{\'o}pez, Antonio and Koltun, Vladlen and Dosovitskiy, Alexey},
  booktitle={2018 IEEE international conference on robotics and automation (ICRA)},
  pages={4693--4700},
  year={2018},
  organization={IEEE}
}

@inproceedings{behaviorclone,
  title={Exploring the limitations of behavior cloning for autonomous driving},
  author={Codevilla, Felipe and Santana, Eder and L{\'o}pez, Antonio M and Gaidon, Adrien},
  booktitle={Proceedings of the IEEE/CVF international conference on computer vision},
  pages={9329--9338},
  year={2019}
}

@inproceedings{roach,
  title={End-to-end urban driving by imitating a reinforcement learning coach},
  author={Zhang, Zhejun and Liniger, Alexander and Dai, Dengxin and Yu, Fisher and Van Gool, Luc},
  booktitle={Proceedings of the IEEE/CVF international conference on computer vision},
  pages={15222--15232},
  year={2021}
}

@inproceedings{st-p3,
  title={St-p3: End-to-end vision-based autonomous driving via spatial-temporal feature learning},
  author={Hu, Shengchao and Chen, Li and Wu, Penghao and Li, Hongyang and Yan, Junchi and Tao, Dacheng},
  booktitle={European Conference on Computer Vision},
  pages={533--549},
  year={2022},
  organization={Springer}
}

@inproceedings{diffusiondrive,
  title={Diffusiondrive: Truncated diffusion model for end-to-end autonomous driving},
  author={Liao, Bencheng and Chen, Shaoyu and Yin, Haoran and Jiang, Bo and Wang, Cheng and Yan, Sixu and Zhang, Xinbang and Li, Xiangyu and Zhang, Ying and Zhang, Qian and others},
  booktitle={Proceedings of the Computer Vision and Pattern Recognition Conference},
  pages={12037--12047},
  year={2025}
}

@inproceedings{cooper,
  title={Cooper: Cooperative perception for connected autonomous vehicles based on 3d point clouds},
  author={Chen, Qi and Tang, Sihai and Yang, Qing and Fu, Song},
  booktitle={2019 IEEE 39th International Conference on Distributed Computing Systems (ICDCS)},
  pages={514--524},
  year={2019},
  organization={IEEE}
}

@inproceedings{who2com,
  title={Who2com: Collaborative perception via learnable handshake communication},
  author={Liu, Yen-Cheng and Tian, Junjiao and Ma, Chih-Yao and Glaser, Nathan and Kuo, Chia-Wen and Kira, Zsolt},
  booktitle={2020 IEEE International Conference on Robotics and Automation (ICRA)},
  pages={6876--6883},
  year={2020},
  organization={IEEE}
}

@article{v2x-sim,
  title={V2X-Sim: Multi-agent collaborative perception dataset and benchmark for autonomous driving},
  author={Li, Yiming and Ma, Dekun and An, Ziyan and Wang, Zixun and Zhong, Yiqi and Chen, Siheng and Feng, Chen},
  journal={IEEE Robotics and Automation Letters},
  volume={7},
  number={4},
  pages={10914--10921},
  year={2022},
  publisher={IEEE}
}

@article{moe-mesh,
  title={Mesh-tensorflow: Deep learning for supercomputers},
  author={Shazeer, Noam and Cheng, Youlong and Parmar, Niki and Tran, Dustin and Vaswani, Ashish and Koanantakool, Penporn and Hawkins, Peter and Lee, HyoukJoong and Hong, Mingsheng and Young, Cliff and others},
  journal={Advances in neural information processing systems},
  volume={31},
  year={2018}
}

@article{gshard,
  title={Gshard: Scaling giant models with conditional computation and automatic sharding},
  author={Lepikhin, Dmitry and Lee, HyoukJoong and Xu, Yuanzhong and Chen, Dehao and Firat, Orhan and Huang, Yanping and Krikun, Maxim and Shazeer, Noam and Chen, Zhifeng},
  journal={arXiv preprint arXiv:2006.16668},
  year={2020}
}

@article{st-moe,
  title={St-moe: Designing stable and transferable sparse expert models},
  author={Zoph, Barret and Bello, Irwan and Kumar, Sameer and Du, Nan and Huang, Yanping and Dean, Jeff and Shazeer, Noam and Fedus, William},
  journal={arXiv preprint arXiv:2202.08906},
  year={2022}
}

@inproceedings{nuscenes,
  title={nuscenes: A multimodal dataset for autonomous driving},
  author={Caesar, Holger and Bankiti, Varun and Lang, Alex H and Vora, Sourabh and Liong, Venice Erin and Xu, Qiang and Krishnan, Anush and Pan, Yu and Baldan, Giancarlo and Beijbom, Oscar},
  booktitle={Proceedings of the IEEE/CVF conference on computer vision and pattern recognition},
  pages={11621--11631},
  year={2020}
}

@inproceedings{coalign,
  title={Robust collaborative 3d object detection in presence of pose errors},
  author={Lu, Yifan and Li, Quanhao and Liu, Baoan and Dianati, Mehrdad and Feng, Chen and Chen, Siheng and Wang, Yanfeng},
  booktitle={2023 IEEE International Conference on Robotics and Automation (ICRA)},
  pages={4812--4818},
  year={2023},
  organization={IEEE}
}

@article{hungarian,
  title={The Hungarian method for the assignment problem},
  author={Kuhn, Harold W},
  journal={Naval research logistics quarterly},
  volume={2},
  number={1-2},
  pages={83--97},
  year={1955},
  publisher={Wiley Online Library}
}
